\documentclass[11pt]{article}

\usepackage{hyperref}
\usepackage{url}
\usepackage{enumitem}
\usepackage{booktabs}   
\usepackage{multirow}   
\usepackage{amsmath} 
\usepackage{xspace}
\usepackage[capitalize,noabbrev]{cleveref}
\usepackage{graphicx}   
\usepackage{subcaption} 
\usepackage{caption}
\usepackage{listings}
\usepackage[most]{tcolorbox}
\usepackage{xcolor} 
\usepackage[ruled,vlined,linesnumbered]{algorithm2e}
\usepackage{wrapfig}
\usepackage{calc}
\usepackage{makecell}
\usepackage[table]{xcolor}
\usepackage{bbm}
\usepackage{amsfonts}

\crefformat{section}{\S#2#1#3}
\crefformat{subsection}{\S#2#1#3}
\crefformat{subsubsection}{\S#2#1#3}
\crefrangeformat{section}{\S#3#1#4 to~\S#5#2#6}
\crefmultiformat{section}{\S#2#1#3}{ and~\S#2#1#3}{, #2#1#3}{ and~#2#1#3}
\Crefformat{figure}{#2Figure~#1#3}
\Crefmultiformat{figure}{Figs.~#2#1#3}{ and~#2#1#3}{, #2#1#3}{ and~#2#1#3}
\Crefformat{table}{#2Table~#1#3}
\Crefmultiformat{table}{Tabs.~#2#1#3}{ and~#2#1#3}{, #2#1#3}{ and~#2#1#3}
\Crefformat{appendix}{#2Appx.~\S#1#3}
\crefformat{algorithm}{Alg.~#2#1#3}
\Crefformat{algorithm}{Alg.~#2#1#3}
\crefname{algocf}{Alg.}{Algs.}
\Crefname{algocf}{Alg.}{Algs.}
\Crefformat{equation}{#2Eq.~#1#3}

\newcommand{\stitle}[1]{\vspace{0.5em}\noindent{\bf #1}}

\usepackage[preprint]{acl}

\usepackage{times}
\usepackage{latexsym}

\usepackage[T1]{fontenc}

\usepackage[utf8]{inputenc}

\usepackage{microtype}

\usepackage{inconsolata}

\usepackage{graphicx}
\newcommand{\sys}{\mbox{\textsc{ALAR}}\xspace}
\usepackage{enumitem}

\title{Adaptive Latent Agentic Reasoning}

\author{
  Dongwon Jung\textsuperscript{1}  \quad 
  Peng Shi\textsuperscript{2}   \quad 
  Yi Zhang\textsuperscript{3}    \quad 
  Junshan Zhang\textsuperscript{1}\quad 
  Muhao Chen\textsuperscript{1} \\
  \\
  \textsuperscript{1}University of California, Davis \quad
  \textsuperscript{2}University of Waterloo \quad
  \textsuperscript{3}Greenshoe, Inc. \\
  \texttt{\{dwojung,jazh,muhchen\}@ucdavis.edu} \;
  \texttt{peng.shi@uwaterloo.ca} \;
  \texttt{yi@greenshoe.ai} 
}

\begin{document}
\maketitle
\begin{abstract}
Large reasoning models improve performance by generating extended chain-of-thought (CoT) reasoning, but this behavior becomes inefficient when applied to LLM agents. Current LLM agents often generate verbose textual reasoning at every decision step and allocate reasoning effort nearly uniformly across turns, leading to substantial inefficiency in multi-turn agentic trajectories. We propose \textbf{Adaptive Latent Agentic Reasoning} (\sys), a dual-mode framework that uses compact latent reasoning for routine turns and selectively escalates to explicit chain-of-thought when deeper deliberation is needed. \sys learns latent reasoning by using the agent's actions as supervision anchors, and is further optimized to use latent reasoning when it is sufficient for task success and reserve explicit CoT for harder decisions. Experiments on agentic search and tool-use benchmarks show that \sys maintains comparable or better task accuracy while substantially reducing generated tokens by up to 43.6\% in search and 84.6\% in tool use. These results demonstrate that \sys improves the accuracy-efficiency trade-off of LLM agents by reducing unnecessary textual reasoning while preserving explicit deliberation for harder decision steps.
\end{abstract}
\section{Introduction}

Recent advances in large reasoning models (LRMs) have shown that extended chain-of-thought (CoT) reasoning improves performance on mathematical, logical, and coding tasks \cite{jaech2024openai,guo2025deepseek}. In the standard single-pass setting, reasoning is primarily answer-directed, where the model deliberates before producing a final response. By contrast, LLM agents extend this paradigm to interactive environments, where reasoning is interleaved with actions such as retrieval, tool use, and environment interaction \cite{yao2022react,shinn2023reflexion}. We refer to this per-turn computation performed at each decision step as agentic reasoning \citep{wei2026agentic}: reasoning used to choose the next action, incorporate observations, and decide when to terminate.

However, current LLM agents largely inherit the reasoning behavior of single-pass LRMs. As a result, they often generate lengthy CoT \cite{chen2024not} even when the next action mainly depends on external observations, and they 
allocate nearly even reasoning effort across turns despite substantial variations in reasoning demands. This inefficiency compounds in multi-turn trajectories, where reasoning tokens from earlier turns accumulate in the growing context. We therefore ask how to make LLM agents reason more efficiently while preserving 
the deliberation needed for challenging decision steps.

\begin{figure}[t]
    \centering
    \small\includegraphics[width=0.95\linewidth]{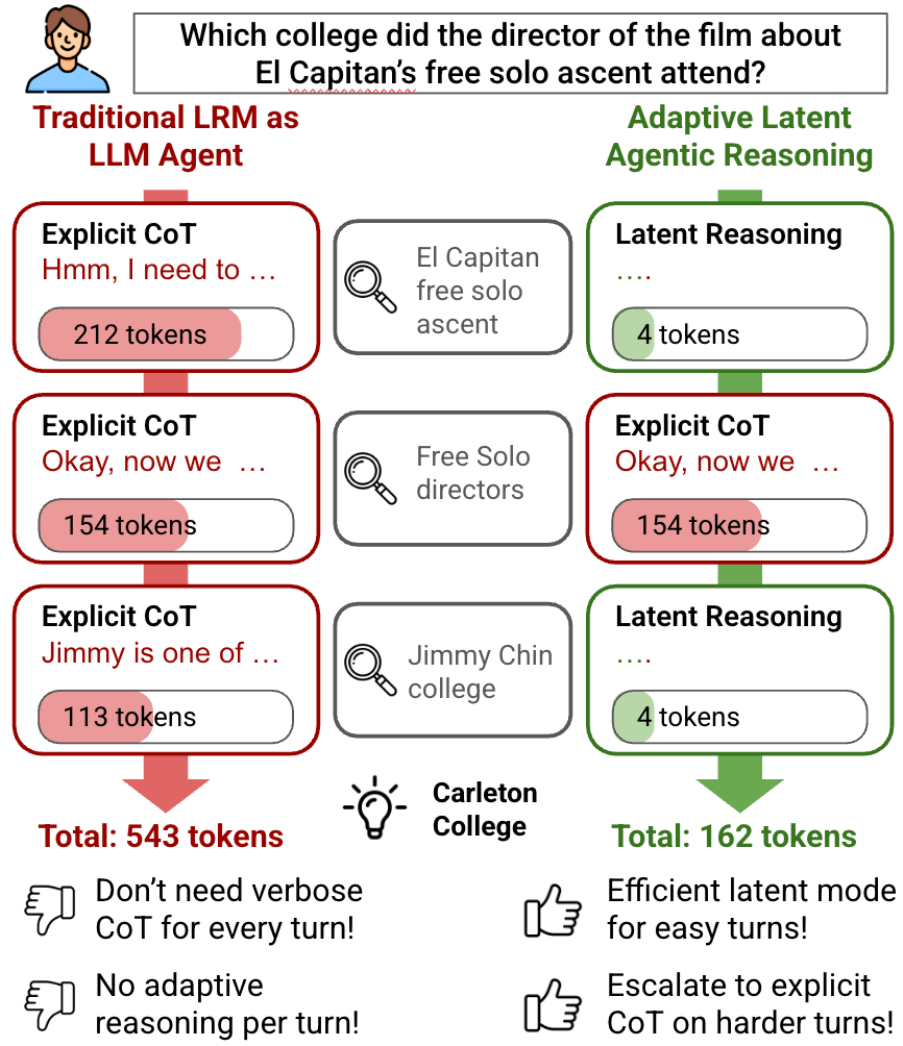}
    \caption{Traditional LRMs generate verbose CoT at every decision step, introducing significant inefficiency in multi-turn agentic trajectories. \sys uses compact latent reasoning by default and falls back to explicit CoT only for turns that require deeper planning.}
    \label{fig:motivation}
    \vspace{-1em}
\end{figure}

A natural approach is to apply recent reasoning token compression methods, which reduce verbose CoT through pruning, length budgets, or rewards for shorter correct solutions \citep{luo2025o1,hou2025thinkprune,yi2026shorterbetter}. However, these methods still operate within the explicit CoT interface where every turn must produce a textual reasoning trace, and efficiency is obtained only by shortening that trace. This is limiting in the agentic setting, where many turns do not require even a shortened textual rationale, but only sufficient internal computation to choose the next environment-coupled action. Thus, efficient agentic reasoning requires a more structural change beyond compressing explicit CoT.

A promising alternative is implicit chain-of-thought or latent reasoning \cite{hao2024training,shen2025codi}, which replaces textual reasoning tokens with a fixed-length sequence of continuous thoughts in the model’s hidden-state space. By avoiding the generation of explicit reasoning tokens, implicit CoT provides a compact form of internal computation. However, extending implicit CoT to agentic settings introduces two challenges. First, training the latent reasoning mode is nontrivial because continuous thoughts live in hidden-state space, so variable-length textual CoT cannot serve as a direct supervision target. Moreover, in agents, per-turn reasoning should support intermediate action selection rather than only final-answer generation. Second, the model should not rely on latent reasoning uniformly. Instead, it must retain the ability to escalate to explicit CoT on turns that genuinely require deeper reasoning, to achieve the desired level of performance.

To address these challenges, we introduce \emph{Adaptive Latent Agentic Reasoning} (\sys), a reasoning architecture for LLM agents that uses latent reasoning by default and escalates to explicit CoT only when the current turn requires deeper reasoning. \sys consists of two components. First, \emph{Action-Anchored Self-Distillation} (AASD) trains the latent reasoning mode without directly supervising latent states. Instead of aligning latent thoughts with textual CoT, AASD replaces each teacher CoT span with a latent block and trains the student to reproduce the teacher's subsequent action. Since actions are the points where the agent interacts with the environment, they provide natural anchors for supervision. Second, \emph{Adaptive Reasoning GRPO} (AR-GRPO) learns adaptive mode selection by rewarding latent reasoning when it preserves task success, while encouraging explicit CoT on turns that require more detailed reasoning.

\begin{figure}[t]
    \centering
    \includegraphics[width=1.0\linewidth]{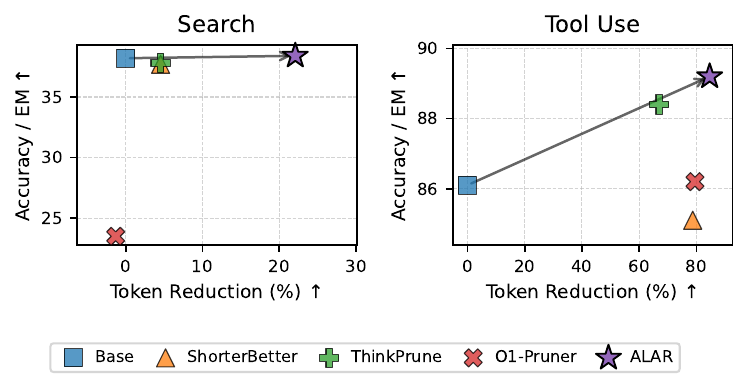}
    \caption{\sys achieves a better accuracy-efficiency trade-off than reasoning token compression baselines across search and tool-use benchmarks.}
    \label{fig:performance}
    \vspace{-1em}
\end{figure}
We evaluate ALAR on agentic search and tool-use benchmarks against recent reasoning token compression baselines. As shown in \Cref{fig:performance}, ALAR achieves a better accuracy-efficiency trade-off by reducing tokens more aggressively while preserving task accuracy. Our contributions are summarized as follows:
\begin{itemize}[leftmargin=1em]
\setlength{\itemsep}{0pt}
    \item We introduce \sys, a dual-mode framework that combines latent reasoning with adaptive mode selection, allowing LLM agents to use compact latent reasoning when it suffices and escalate to explicit CoT 
    at turns where additional deliberation is needed for action selection.
    \item We propose \emph{Action-Anchored Self-Distillation} (AASD), a self-distillation method 
    which trains latent agentic reasoning without latent-state supervision by replacing teacher CoT spans with latent blocks and supervising the student to reproduce the teacher’s next environment-facing action.
    \item We propose \emph{AR-GRPO}, a reinforcement learning method that 
    optimizes per-turn reasoning-mode selection by rewarding latent-mode use when task success is preserved and discouraging unnecessary explicit CoT
\end{itemize}

\begin{figure*}[t]
    \centering
    \small\includegraphics[width=0.9\linewidth]{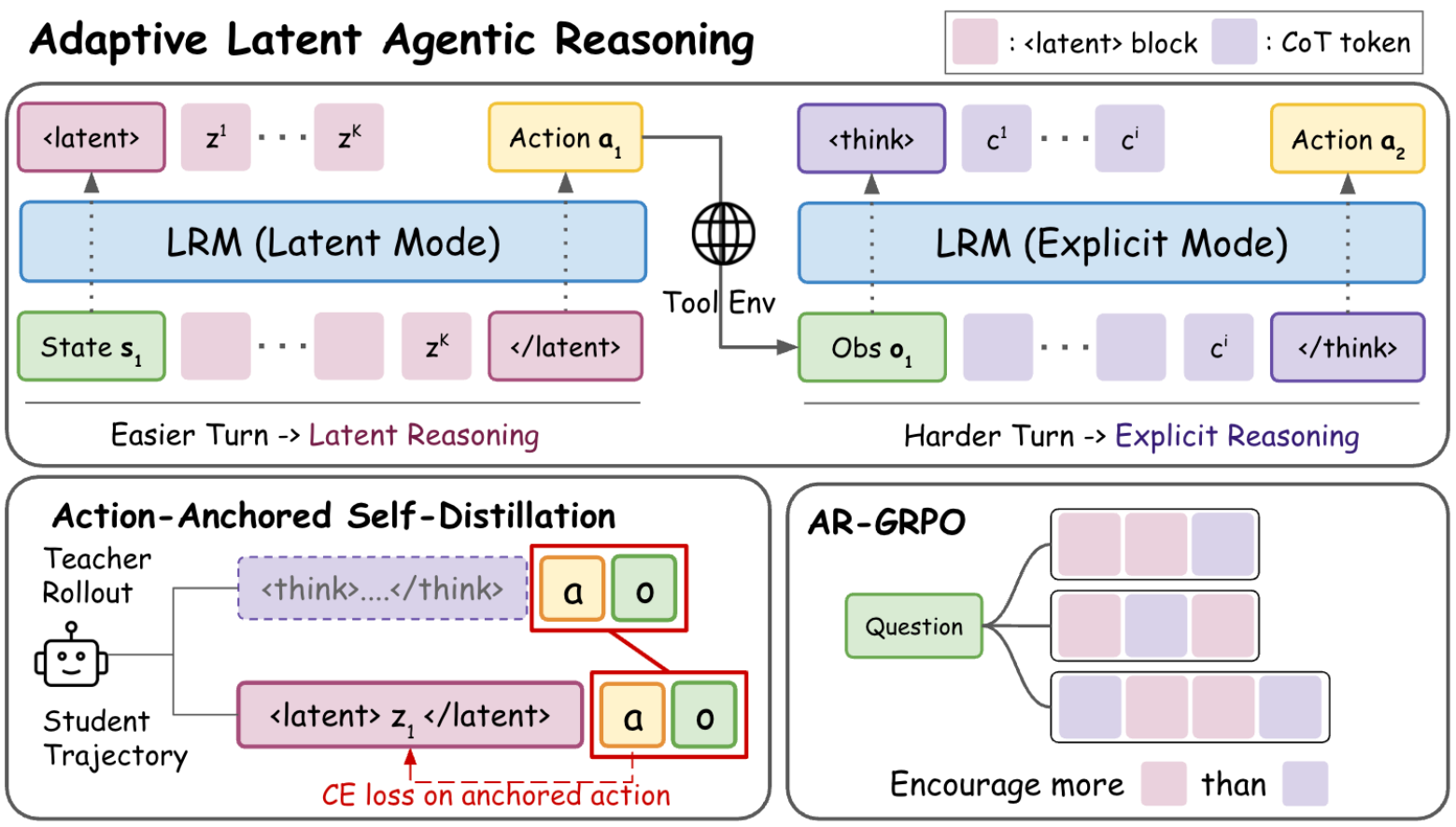}
    \caption{Overview of \sys. At each turn, LRM adaptively chooses latent mode for routine decisions or explicit mode for harder turns. Action-Anchored Self-Distillation trains the latent mode by using the teacher's actions as anchors. AR-GRPO further learns when to use latent reasoning by rewarding it only when task success is preserved.}
    \label{fig:main}
    \vspace{-1em}
\end{figure*}
\section{Related Work}
\subsection{Latent Reasoning}
Recent work has explored latent reasoning as an efficient alternative to explicit CoT. Early methods train models to internalize or compress textual CoT into continuous hidden states \citep{deng2024explicit,hao2024training, shen2025codi,cheng2024compressed}. More recent hybrid approaches combine latent and explicit reasoning through switching, gating, or token-level mixing \citep{shi2025swireasoning,xu2026thinkrouter,yue2026hybrid,su2025token}. These methods mainly target single-pass reasoning, where latent computation is used to produce a final answer. Our setting differs in that latent reasoning is action-oriented, environment-coupled, and repeated across turns, making the central challenge not only how to compress reasoning, but also how to allocate reasoning modes throughout a trajectory.

\subsection{Reasoning Token Reduction}
To mitigate overthinking in LRMs \cite{chen2024not,sui2025stop}, recent work has sought to reduce reasoning cost by shortening explicit CoT traces. One group of methods uses reinforcement learning or fine-tuning rewards to favor concise-but-correct reasoning and prune redundant thinking steps \citep{arora2026training, luo2025o1, hou2025thinkprune, cheng2025optimizing}. Another group introduces length control or difficulty-adaptive budgets, allowing models to adjust reasoning length according to a user-specified budget, sampled optimal length, or problem difficulty \citep{aggarwal2025l1, yi2026shorterbetter, shen2025dast}. While effective, these methods still optimize efficiency within the textual CoT interface. Our work instead changes the reasoning substrate itself, using latent reasoning to bypass unnecessary textual CoT and enable more aggressive token reduction across multi-turn trajectories.
\section{Adaptive Latent Agentic Reasoning}
\label{sec:method}

To this end, we propose \textbf{Adaptive Latent Agentic Reasoning} (\sys), a dual-mode reasoning framework for efficient LLM agents. We first formulate the LRM as a multi-turn agent policy (\Cref{sec:lrm_as_llm_agents}), then introduce two core design components: \emph{Latent Agentic Reasoning} (\Cref{sec:latent_agentic_reasoning}) and \emph{Adaptive Mode Selection} (\Cref{sec:adaptive_mode_selection}). We then present the two-stage optimization procedure: \emph{Action-Anchored Self-Distillation} (\Cref{sec:aasd}) learns the latent agentic reasoning, and AR-GRPO learns adaptive mode selection (\Cref{sec:ar_grpo}).

\subsection{LRMs as LLM Agents}
\label{sec:lrm_as_llm_agents}

We consider a large reasoning model (LRM) parameterized by $\theta$ that produces an explicit chain-of-thought (CoT) before each output \citep{guo2025deepseek,xiang2025towards}. 
In an agentic setting, this reason-before-output pattern is repeated across multiple environment-coupled decision steps. Specifically, we treat the LRM as the policy of an LLM agent that interacts with a tool environment over up to $T$ turns. Given a query $x$, at each turn $t$ the agent generates an explicit CoT $c_t \sim \pi_\theta(\cdot \mid s_t)$, conditioned on the state $s_t$ (the current context), then emits an action $a_t \sim \pi_\theta(\cdot \mid s_t, c_t)$ that is either a tool call or the final response. If $a_t$ is a tool call, the environment returns an observation $o_t$ that is appended to the context; otherwise, the episode terminates. The resulting trajectory is $\tau = (x, c_1, a_1, o_1, \ldots, c_T, a_T)$, where $a_T$ is the final response.

\subsection{Latent Agentic Reasoning}
\label{sec:latent_agentic_reasoning}
The formulation exposes the main inefficiency we target: explicit CoT is generated at every turn, even when the next action may require only lightweight internal computation.
Latent reasoning has so far been studied primarily in single-pass reasoning tasks \citep{hao2024training, shen2025codi}, where continuous thoughts replace the CoT before producing a final answer. We adapt this idea to multi-turn agentic reasoning, where reasoning serves a different role: at each intermediate turn, the agent reasons to select the next action toward a long-horizon goal rather than to directly produce the final answer.

Specifically, at each turn $t$, instead of generating an explicit CoT $c_t$, the agent produces a fixed-length sequence of $K$ continuous thoughts $z_t = (z_t^1, \ldots, z_t^K)$. Starting from the hidden state $h_t^0$ corresponding to the current state $s_t$, each latent thought is generated autoregressively in hidden-state space:
\[
z_t^k = f_\phi(h_t^{k-1}), \quad k = 1, \ldots, K,
\]
where $f_\phi$ is a projection layer and each $z_t^k$ is fed back as the input embedding for the next latent position. After the latent block is produced, the agent samples the next action as $a_t \sim \pi_\theta(\cdot \mid s_t, z_t)$, where $a_t$ is decoded over the vocabulary $V$ conditioned on the current state and the latent thoughts. We refer to this process as latent agentic reasoning: the agent performs implicit per-turn computation through a latent block rather than a discrete CoT.

\subsection{Adaptive Mode Selection}
\label{sec:adaptive_mode_selection}
Although the latent agentic reasoning is sufficient for routine turns, some decisions require more substantive reasoning than a fixed-length latent block can accommodate. We therefore equip the agent with a per-turn choice between latent and explicit \emph{mode}, with the mode sampled directly from the policy:
\begin{equation*}
m_t \sim \pi_\theta(\cdot \mid s_t), \qquad r_t \sim \pi_\theta(\cdot \mid s_t, m_t),
\label{eq:mode}
\end{equation*}
where the mode $m_t \in \{\textsc{<lat>}, \textsc{<think>}\}$ determines the form of the per-turn reasoning trace $r_t$, which is the latent block $z_t$ in the latent mode and an explicit CoT $c_t$ in the explicit mode. The action is then sampled from $\pi_\theta(\cdot \mid s_t, r_t)$.
Letting $r_t \in \{z_t, c_t\}$ denote the reasoning trace of turn $t$ under its selected mode, the resulting trajectory is $\tau = (x, m_1, r_1, a_1, o_1, \ldots, m_T, r_T, a_T)$. Because $m_t$ is sampled from the same policy that generates the rest of the trajectory, mode selection becomes part of the agent's decision space rather than a choice imposed by an external orchestrator or router.



\subsection{Action-Anchored Self-Distillation}
\label{sec:aasd}
  
Training the latent mode raises a supervision challenge. The projector $f_\phi$ that produces the continuous thoughts $z_t$ is newly initialized and has no targets to learn from. An obvious candidate is the explicit CoT $c_t$ that $z_t$ replaces, but the two are structurally mismatched: $c_t$ is a variable-length sequence of discrete tokens, whereas $z_t$ is a fixed-length sequence of continuous vectors. Matching them position-wise would tie $f_\phi$ to the token-level decomposition of the teacher's reasoning instead of letting it discover its own.

We address this with Action-Anchored Self-Distillation (AASD): the same base model acts as a teacher in the explicit mode and a student in the latent mode, with the student anchored to the teacher's \emph{actions}. Anchoring on actions sidesteps the alignment problem: actions are the points at which both modes contact the environment and at which correctness is defined, so $f_\phi$ is free to discover whatever trajectory through hidden-state space best produces $a_t$ from $s_t$, without being told what $z_t$ should look like.

\stitle{Teacher rollouts.} Let $\pi_{\theta}$ denote the base LRM, shared between the two modes. We roll out the explicit mode on a training set in the agentic environment, and from each resulting trajectory we extract the \emph{action trajectory}
$\tau_a = (a_1, o_1, a_2, o_2, \ldots, a_{T-1}, o_{T-1}, a_T)$,
which retains the teacher's actions and the corresponding environment observations while dropping its explicit CoTs. 

\stitle{Student objective.} The student shares the base parameters $\theta$ with the teacher and operates in the latent mode, with the projector $f_\phi$ providing the continuous thoughts. Given an action trajectory $\tau_a$, we form a student trajectory
$\tilde{\tau} = (x, m_1, z_1, a_1, o_1, \ldots, m_T, z_T, a_T)$
by inserting a latent mode token $m_t=\textsc{<lat>}$ and a latent block $z_t$ of length $K$ before each anchor action $a_t$ in place of the teacher's CoT. 

We train $(\theta, \phi)$ by maximizing the log-likelihood of the teacher's anchor actions under the student trajectory, conditioned on the state $s_t$ and the preceding latent block $z_t$, produced by the projector chain $z_t^k = f_\phi(h_t^{k-1})$:
\begin{equation*}
\begin{aligned}
\mathcal{L}_{\text{AASD}}
= -\mathbb{E}_{\tau_a}\sum_{t=1}^{T}
\big[
&\log \pi_\theta(m_t \mid s_t) \\
&+ \log \pi_\theta(a_t \mid s_t,m_t,z_t)
\big].
\end{aligned}
\label{eq:aasd-loss}
\end{equation*}
The loss is applied to the action tokens and the mode tokens $m_t$, so that the model also learns to emit the mode tokens at the start of each turn. The $K$ latent positions have no discrete token target to compute cross-entropy against, since $z_t$ lives in continuous space rather than over the vocabulary $V$, and the environment observations $o_t$ are masked out to stabilize training \citep{jin2025search}. The latent block $z_t$ that the student inserts between $s_t$ and $a_t$ is learned end-to-end: the cross-entropy at each anchor action $a_t$ back-propagates through the transformer to the $K$ latent input positions and from there through the iterative projector chain $z_t^k = f_\phi(h_t^{k-1})$, accumulating gradient contributions across all $K$ projector steps.

\subsection{AR-GRPO}
\label{sec:ar_grpo}

After learning the latent mode with AASD, we train adaptive mode selection by first initializing the mode distribution with a brief mode-warmup SFT and then optimizing the policy with AR-GRPO. The goal is to encourage latent reasoning whenever it improves efficiency without sacrificing task success, while preserving the ability to escalate to explicit CoT when needed.

\stitle{Mode warmup.}
AASD trains every turn with \textsc{<LAT>}, so the resulting policy has little probability mass on \texttt{<THINK>} and provides weak exploration for adaptive mode selection. We therefore begin with a brief mode-warmup SFT: starting from the AASD checkpoint, we assign each turn in a small subset of teacher trajectories to either \textsc{<LAT>} or \texttt{<THINK>}. \textsc{<LAT>} turns are trained with the AASD objective, while \texttt{<THINK>} turns are trained with standard cross-entropy on the teacher's CoT. 

\stitle{Trajectory reward.}
After warmup, we optimize the agent over complete trajectories. For each query, we sample a group of $G$ rollouts $\{\tau^{(i)}\}_{i=1}^{G}$ from $\pi_\theta$. Let $n_{\mathrm{LAT}}(\tau)$ denote the number of latent reasoning turns and $n_{\mathrm{turn}}(\tau)$ denote the total number of reasoning turns. We define the \emph{latent fraction} of a trajectory as
\[
f(\tau)=
\frac{n_{\mathrm{LAT}}(\tau)}
{n_{\mathrm{turn}}(\tau)}
\in[0,1],
\]
with $f(\tau)=0$ when no reasoning turn is taken.

Based on this latent fraction, we define an asymmetric format reward that encourages latent reasoning only when it preserves task success:
\[
r_{\mathrm{fmt}}(\tau)=
\begin{cases}
1+\alpha f(\tau), & \mathrm{EM}(\tau)=1,\\
-\alpha f(\tau), & \mathrm{otherwise},
\end{cases}
\]
where $\alpha>0$ controls the latent mode bonus. Intuitively, correct trajectories are rewarded more when they rely more on latent reasoning, while incorrect trajectories are penalized for overusing latent reasoning.

To avoid early collapse to a single mode mixture, we add a decayed diversity bonus,
\[
r_{\mathrm{div}}(\tau^{(i)}) =
d_s \left| f(\tau^{(i)})-\bar{f}_G \right|,
\]
where $\bar{f}_G$ denote the group mean latent fraction and $d_s$ cosine-decays from $1$ to $0$ during training. This term encourages early exploration of different latent-explicit mixtures, then fades so that the success-conditioned format reward dominates.

Finally, we apply a length-scaling factor $s_L(\tau)$ that remains $1$ within the tolerance length $L$ and down-weights trajectories with overlong explicit \texttt{<THINK>} segments. The final trajectory reward combines the format and diversity terms under this length scaling:
\[
R^{(i)} =
s_L(\tau^{(i)})
\left(
r_{\mathrm{fmt}}(\tau^{(i)})+
r_{\mathrm{div}}(\tau^{(i)})
\right),
\]
with $R^{(i)}=-1$ for invalid output formats.

\stitle{GRPO optimization.}
We normalize the trajectory rewards within each rollout group to obtain the advantage $\hat{A}^{(i)}=(R^{(i)}-\mu)/\sigma$, where $\mu$ and $\sigma$ are the mean and standard deviation of $\{R^{(j)}\}_{j=1}^{G}$. This advantage is broadcast to all policy-generated tokens in $\tau^{(i)}$, and $\pi_\theta$ is optimized with the standard GRPO clipped objective with a KL penalty to the reference policy \citep{shao2024deepseekmath}.

\section{Experiment Setting}

We evaluate \sys in two agentic domains, \emph{search} and \emph{tool use}. We first describe the implementation details on both domains and then illustrate the evaluation setup.

\subsection{Implementation}
\label{sec:implementation}

\textbf{Models and Datasets.}
In the search domain, we use the released Search-R1 \cite{jin2025search} 3B and 7B checkpoints as the base LRM, which are RL-trained on NQ \cite{kwiatkowski2019natural} and HotpotQA \cite{yang2018hotpotqa}. We roll out each base model in explicit mode on its training pool and keep only successful trajectories using exact-matching rejection sampling, yielding 86K trajectories for the 7B model and 76K for the 3B model. 

In the tool-use domain, we use Qwen3-4B-Thinking \cite{yang2025qwen3}, a 4B LRM with native tool-calling capability. Given a query and a set of candidate tools in the system prompt, the model emits a multi-step tool-calling trajectory in a single assistant turn, interleaving reasoning with JSON function calls. Teacher trajectories are collected from the \texttt{graph\_syn} subset of ToolMind \cite{yang2025toolmind}, and we retain only rollouts whose tool calls exactly match the reference calls under AST-level matching, resulting in 21K teacher rollouts. For AR-GRPO, we use $G=8$, $\alpha=0.3$ in both domains and set generous generation length tolerances of $L=400$ for search and $L=1600$ for tool use.

\begin{table*}[t]
    \centering
    \scriptsize
    \setlength{\tabcolsep}{3pt}
    \begin{tabular}{ll *{7}{ccc}}
    \toprule
    \multicolumn{2}{c}{}
    & \multicolumn{3}{c}{NQ}
    & \multicolumn{3}{c}{HotpotQA}
    & \multicolumn{3}{c}{TriviaQA}
    & \multicolumn{3}{c}{2Wiki}
    & \multicolumn{3}{c}{MuSiQue}
    & \multicolumn{3}{c}{Bamboogle}
    & \multicolumn{3}{c}{Avg.} \\
    \cmidrule(lr){3-5}
    \cmidrule(lr){6-8}
    \cmidrule(lr){9-11}
    \cmidrule(lr){12-14}
    \cmidrule(lr){15-17}
    \cmidrule(lr){18-20}
    \cmidrule(lr){21-23}
    \multicolumn{2}{c}{Method}
    & EM & Tok & $\mathrm{AE}$
    & EM & Tok & $\mathrm{AE}$
    & EM & Tok & $\mathrm{AE}$
    & EM & Tok & $\mathrm{AE}$
    & EM & Tok & $\mathrm{AE}$
    & EM & Tok & $\mathrm{AE}$
    & EM & Tok & $\mathrm{AE}$ \\
    \midrule
    \multicolumn{23}{l}{\textit{Qwen2.5-3B}} \\
    \multicolumn{2}{l}{Search-R1} & 42.9 & 138 & \textcolor{gray}{0.00} & 37.4 & 158 & \textcolor{gray}{0.00} & 61.3 & 143 & \textcolor{gray}{0.00} & 39.6 & 172 & \textcolor{gray}{0.00} & 14.6 & 174 & \textcolor{gray}{0.00} & 33.6 & 141 & \textcolor{gray}{0.00} & 38.2 & 154 & \textcolor{gray}{0.00} \\
    \multicolumn{2}{l}{ShorterBetter} & 41.3 & 132 & \textcolor{red!56}{-0.14} & 36.9 & 150 & \textcolor{red!42}{-0.02} & 60.2 & 137 & \textcolor{red!45}{-0.05} & \textbf{39.0} & 166 & \textcolor{red!44}{-0.04} & \textbf{15.3} & 168 & \textcolor{green!45!black!60}{+0.18} & 33.6 & 129 & \textcolor{green!45!black!49}{+0.09} & 37.7 & 147 & \textcolor{gray}{0.00} \\
    \multicolumn{2}{l}{ThinkPrune} & 41.2 & 132 & \textcolor{red!57}{-0.15} & 37.0 & 150 & \textcolor{gray}{0.00} & \textbf{60.3} & 137 & \textcolor{red!44}{-0.04} & 38.8 & 166 & \textcolor{red!47}{-0.07} & \textbf{15.3} & 168 & \textcolor{green!45!black!60}{+0.18} & 34.4 & 130 & \textcolor{green!45!black!56}{+0.15} & 37.8 & 147 & \textcolor{green!45!black!41}{+0.01} \\
    \multicolumn{2}{l}{O1-Pruner} & 19.3 & 149 & \textcolor{red!90}{-2.83} & 19.8 & 158 & \textcolor{red!90}{-2.35} & 41.7 & 155 & \textcolor{red!90}{-1.68} & 26.1 & 168 & \textcolor{red!90}{-1.68} & 5.2 & 167 & \textcolor{red!90}{-3.18} & 28.8 & 137 & \textcolor{red!90}{-0.69} & 23.5 & 156 & \textcolor{red!90}{-2.07} \\
    \midrule
    \multirow{2}{*}{\textbf{ALAR}} & Stage 1 & \textbf{41.4} & \textbf{74} & \textbf{\textcolor{green!45!black!72}{+0.29}} & \textbf{38.0} & \textbf{91} & \textbf{\textcolor{green!45!black!90}{+0.47}} & 55.9 & \textbf{113} & \textcolor{red!65}{-0.23} & 38.5 & \textbf{94} & \textbf{\textcolor{green!45!black!75}{+0.31}} & 14.8 & \textbf{105} & \textbf{\textcolor{green!45!black!88}{+0.44}} & 35.2 & \textbf{87} & \textbf{\textcolor{green!45!black!90}{+0.53}} & 37.3 & \textbf{94} & \textbf{\textcolor{green!45!black!73}{+0.30}} \\
     & Stage 2 & 41.3 & 106 & \textcolor{green!45!black!45}{+0.05} & \textbf{38.0} & 125 & \textcolor{green!45!black!68}{+0.26} & 60.1 & 115 & \textbf{\textcolor{green!45!black!51}{+0.10}} & \textbf{39.0} & 130 & \textcolor{green!45!black!59}{+0.17} & \textbf{15.3} & 138 & \textcolor{green!45!black!79}{+0.35} & \textbf{36.8} & 106 & \textbf{\textcolor{green!45!black!90}{+0.53}} & \textbf{38.4} & 120 & \textcolor{green!45!black!67}{+0.24} \\
    \midrule
    \multicolumn{23}{l}{\textit{Qwen2.5-7B}} \\
    \multicolumn{2}{l}{Search-R1} & 49.1 & 205 & \textcolor{gray}{0.00} & 43.2 & 250 & \textcolor{gray}{0.00} & 63.8 & 234 & \textcolor{gray}{0.00} & 40.1 & 257 & \textcolor{gray}{0.00} & 19.1 & 249 & \textcolor{gray}{0.00} & 40.8 & 218 & \textcolor{gray}{0.00} & 42.7 & 236 & \textcolor{gray}{0.00} \\
    \multicolumn{2}{l}{ShorterBetter} & 46.6 & 190 & \textcolor{red!60}{-0.18} & \textbf{42.8} & 232 & \textcolor{green!45!black!43}{+0.03} & \textbf{64.4} & 215 & \textcolor{green!45!black!52}{+0.11} & 41.3 & 233 & \textcolor{green!45!black!60}{+0.18} & 18.5 & 231 & \textcolor{red!49}{-0.08} & 37.6 & 196 & \textcolor{red!72}{-0.29} & 41.9 & 216 & \textcolor{red!44}{-0.04} \\
    \multicolumn{2}{l}{ThinkPrune} & \textbf{46.8} & 191 & \textcolor{red!58}{-0.17} & \textbf{42.8} & 236 & \textcolor{green!45!black!41}{+0.01} & 64.3 & 218 & \textcolor{green!45!black!50}{+0.09} & \textbf{42.0} & 238 & \textcolor{green!45!black!64}{+0.22} & 18.4 & 234 & \textcolor{red!54}{-0.12} & 37.6 & 200 & \textcolor{red!74}{-0.31} & \textbf{42.0} & 220 & \textcolor{red!45}{-0.05} \\
    \multicolumn{2}{l}{O1-Pruner} & \textbf{46.8} & 190 & \textcolor{red!58}{-0.16} & \textbf{42.8} & 227 & \textcolor{green!45!black!45}{+0.05} & 64.3 & 213 & \textcolor{green!45!black!52}{+0.11} & 41.0 & 228 & \textcolor{green!45!black!60}{+0.18} & \textbf{18.8} & 228 & \textcolor{green!45!black!41}{+0.01} & 36.8 & 192 & \textcolor{red!81}{-0.37} & 41.8 & 213 & \textcolor{red!43}{-0.03} \\
    \midrule
    \multirow{2}{*}{\textbf{ALAR}} & Stage 1 & 46.3 & \textbf{106} & \textbf{\textcolor{green!45!black!62}{+0.20}} & 42.0 & \textbf{112} & \textbf{\textcolor{green!45!black!85}{+0.41}} & 63.5 & \textbf{112} & \textbf{\textcolor{green!45!black!90}{+0.50}} & 38.8 & \textbf{129} & \textcolor{green!45!black!77}{+0.34} & 17.7 & \textbf{115} & \textcolor{green!45!black!59}{+0.17} & 34.4 & \textbf{97} & \textcolor{red!65}{-0.23} & 40.5 & \textbf{112} & \textcolor{green!45!black!65}{+0.23} \\
     & Stage 2 & \textbf{46.8} & 123 & \textcolor{green!45!black!58}{+0.17} & \textbf{42.8} & 140 & \textcolor{green!45!black!83}{+0.39} & 64.3 & 133 & \textcolor{green!45!black!90}{+0.46} & 39.6 & 144 & \textbf{\textcolor{green!45!black!82}{+0.38}} & 18.4 & 135 & \textbf{\textcolor{green!45!black!70}{+0.27}} & \textbf{38.4} & 124 & \textbf{\textcolor{green!45!black!55}{+0.14}} & 41.7 & 133 & \textbf{\textcolor{green!45!black!73}{+0.30}} \\
    \bottomrule
    \end{tabular}
    \caption{Evaluation results on the search domain using Search-R1 as the base model, at the Qwen2.5-3B and Qwen2.5-7B scales.}
    \label{tab:search_result}
    \vspace{-0.5em}
\end{table*}
\begin{table*}[t]
    \centering
    \scriptsize
    \setlength{\tabcolsep}{5pt}
    \begin{tabular}{ll *{5}{ccc}}
    \toprule
    \multicolumn{2}{c}{}
    & \multicolumn{3}{c}{Simple}
    & \multicolumn{3}{c}{Multiple}
    & \multicolumn{3}{c}{Parallel}
    & \multicolumn{3}{c}{Par.-Mult.}
    & \multicolumn{3}{c}{Avg.}
    \\
    \cmidrule(lr){3-5}
    \cmidrule(lr){6-8}
    \cmidrule(lr){9-11}
    \cmidrule(lr){12-14}
    \cmidrule(lr){15-17}
    \multicolumn{2}{c}{Method}
    & Acc & Tok & $\mathrm{AE}$
    & Acc & Tok & $\mathrm{AE}$
    & Acc & Tok & $\mathrm{AE}$
    & Acc & Tok & $\mathrm{AE}$
    & Acc & Tok & $\mathrm{AE}$
    \\
    \midrule
    \multicolumn{17}{l}{\textit{Qwen3-4B-Thinking}} \\
    \multicolumn{2}{l}{Qwen3-4B} & 91.5 & 564 & \textcolor{gray}{0.00} & 92.0 & 501 & \textcolor{gray}{0.00} & 85.0 & 891 & \textcolor{gray}{0.00} & 76.0 & 1083 & \textcolor{gray}{0.00} & 86.1 & 760 & \textcolor{gray}{0.00} \\
    \multicolumn{2}{l}{ShorterBetter} & 93.0 & 109 & \textcolor{green!45!black!90}{+0.86} & 90.0 & 104 & \textcolor{green!45!black!90}{+0.68} & 82.0 & 201 & \textcolor{green!45!black!90}{+0.60} & 75.5 & 233 & \textcolor{green!45!black!90}{+0.75} & 85.1 & 162 & \textcolor{green!45!black!90}{+0.72} \\
    \multicolumn{2}{l}{ThinkPrune} & 93.5 & 169 & \textcolor{green!45!black!90}{+0.77} & 89.5 & 159 & \textcolor{green!45!black!90}{+0.55} & 89.5 & 325 & \textcolor{green!45!black!90}{+0.79} & 81.0 & 349 & \textcolor{green!45!black!90}{+0.87} & 88.4 & 251 & \textcolor{green!45!black!90}{+0.75} \\
    \multicolumn{2}{l}{O1-Pruner} & 92.5 & 100 & \textcolor{green!45!black!90}{+0.85} & 89.5 & 99 & \textcolor{green!45!black!90}{+0.67} & 87.5 & 195 & \textcolor{green!45!black!90}{+0.87} & 75.5 & 230 & \textcolor{green!45!black!90}{+0.75} & 86.2 & 156 & \textcolor{green!45!black!90}{+0.79} \\
    \midrule
    \multirow{2}{*}{\textbf{ALAR}} & Stage 1 & 93.8 & \textbf{51} & \textbf{\textcolor{green!45!black!90}{+0.98}} & 90.0 & \textbf{50} & \textbf{\textcolor{green!45!black!90}{+0.79}} & \textbf{90.0} & \textbf{116} & \textbf{\textcolor{green!45!black!90}{+1.05}} & 81.5 & \textbf{130} & \textcolor{green!45!black!90}{+1.10} & 88.8 & \textbf{87} & \textbf{\textcolor{green!45!black!90}{+0.98}} \\
     & Stage 2 & \textbf{94.2} & 87 & \textcolor{green!45!black!90}{+0.94} & \textbf{90.5} & 84 & \textcolor{green!45!black!90}{+0.75} & 89.5 & 145 & \textcolor{green!45!black!90}{+1.00} & \textbf{82.5} & 152 & \textbf{\textcolor{green!45!black!90}{+1.12}} & \textbf{89.2} & 117 & \textcolor{green!45!black!90}{+0.95} \\
    \bottomrule
    \end{tabular}
    \caption{Evaluation results on the BFCL benchmark using Qwen3-4B-Thinking as the base model.}
    \label{tab:tool_result}
    \vspace{-1em}
\end{table*}

\stitle{Latent block.}
Each latent block consists of $K=4$ continuous thoughts framed by surface tags \textsc{<LAT>}...\textsc{</LAT>}. The four placeholders are repurposed as content-free sentinels: at every \textsc{<LAT>}, the projector $f_\phi$ writes $K$ continuous embeddings into these positions, and the closing \textsc{</LAT>} is prefilled after the $K$ projections programmatically. The tags are standard tokens in the vocabulary of the model, so no new special token is added. The projector is a two-layer MLP with GELU and a final LayerNorm whose hidden width matches the base model.

\stitle{Training.}
Both domains follow the same two-stage training pipeline. Stage~1 trains the latent mode with AASD on successful teacher trajectories, replacing each teacher reasoning span with \textsc{<LAT>} followed by a length-$K$ latent block and supervising only the subsequent anchor actions. For the mode warmup, we first perform a brief SFT from the Stage~1 checkpoint using 20K instances, where each turn is randomly assigned to either latent or explicit thinking with equal probability. We then optimize it with the AR-GRPO objective as the Stage 2.

\subsection{Evaluation Setup}

\stitle{Benchmarks.}
For the search domain, we evaluate on six open-domain QA benchmarks: NQ, HotpotQA, TriviaQA \cite{joshi2017triviaqa}, 2WikiMultiHopQA (2Wiki; \citealt{ho2020constructing}), MuSiQue \cite{trivedi2022musique}, and Bamboogle \cite{press2023measuring}. For the tool-use domain, we evaluate on the AST-based BFCL \cite{patil2025berkeley} on all categories: simple, multiple, parallel, and parallel-multiple. 

\stitle{Evaluation metrics.}
We report task accuracy ($\mathrm{EM}$), average number of generated tokens ($\mathrm{Tok}$), and an Accuracy-Efficiency ($\mathrm{AE}$) score following \citet{luo2025o1}. $\mathrm{EM}$ is exact-match accuracy for search and AST-level tool-call matching for tool use. $\mathrm{Tok}$ counts the model-generated tokens, including reasoning traces, mode tags, tool calls, and final answers. $\mathrm{AE}$ summarizes the accuracy-efficiency trade-off relative to the corresponding base model. Specifically, we compute $\mathrm{AE}=\alpha\Delta_{\mathrm{Tok}}+\beta[\Delta_{\mathrm{EM}}]_+ + \gamma[\Delta_{\mathrm{EM}}]_-$, where $\Delta_{\mathrm{Tok}}=(\mathrm{Tok}_0-\mathrm{Tok})/\mathrm{Tok}_0$, $\Delta_{\mathrm{EM}}=(\mathrm{EM}-\mathrm{EM}_0)/\mathrm{EM}_0$, $[x]_+=\max(0,x)$, and $[x]_-=\min(0,x)$. Here, $\mathrm{EM}_0$ and $\mathrm{Tok}_0$ denote the accuracy and length of the corresponding base model. Following \citet{luo2025o1}, we set $(\alpha,\beta,\gamma)=(1,3,5)$ to penalize accuracy degradation more strongly than accuracy improvement.

\stitle{Baselines.}
In each domain, we compare \sys against three published reasoning-token compression methods: \emph{O1-Pruner} \citep{luo2025o1}, which rewards concise rollouts relative to a reference baseline; \emph{ThinkPrune} \citep{hou2025thinkprune}, which enforces annealed length budgets on thinking spans; and \emph{ShorterBetter} \citep{yi2026shorterbetter}, which encourages rollouts to match the shortest correct reasoning length in each group.

\section{Experiment Results}
\label{sec:results}

\subsection{Main Results}

\Cref{tab:search_result} and \Cref{tab:tool_result} report results on the search and tool-use domains. Overall, \sys achieves the best accuracy--efficiency trade-off across both domains: it matches or improves the base model's EM while substantially reducing generated tokens, yielding the strongest AE Pareto performance.

\stitle{\sys achieves strong token reduction while preserving accuracy.}
In the search domain, \sys substantially reduces generation with little or no accuracy loss. For 3B, Stage 2 improves average EM from $38.2$ to $38.4$ while reducing generated tokens by $22.1\%$. Stage~1 is even more efficient, reducing tokens by $39.0\%$ with competitive EM. For 7B, Stage 2 reduces tokens by $43.6\%$ while maintaining comparable EM, and Stage 1 achieves a $52.5\%$ token reduction.

\stitle{Text-based reasoning compression has limited headroom for search agents.}
The reasoning token reduction baselines provide only modest gains in the search domain. Since Search-R1 already produces compact explicit CoT, methods that only shorten textual reasoning reduce average tokens by about $4$--$10\%$. In contrast, \sys changes the reasoning interface itself by replacing textual reasoning with latent reasoning, enabling much larger reductions without severe performance degradation.

\stitle{\sys is especially effective in the tool-use domain.}
In tool use, the base Qwen3-4B-Thinking model is much more verbose than Search-R1. All compression baselines therefore achieve substantial token reductions, but \sys performs best. Stage~1 improves average accuracy from $86.1$ to $88.8$ while reducing generated tokens by $88.6\%$. Stage 2 further improves accuracy to $89.2$ while reducing tokens by $84.6\%$, achieving the best EM and the strongest AE score among all methods.

\stitle{Stage 1 shows the strength of AASD, while Stage 2 improves adaptivity.}
Stage 1 is highly competitive despite using the fewest tokens, showing that AASD effectively injects latent reasoning into agentic policies. Since Stage 1 uses latent reasoning for every turn, its strong performance suggests that many agentic decisions do not require explicit CoT. Stage 2 generally uses more tokens but improves EM by learning adaptive mode selection through AR-GRPO, allowing the model to use latent reasoning for easier turns and explicit CoT for harder ones.

\stitle{Overall, adaptive latent reasoning outperforms explicit reasoning compression.}
These results support two hypotheses behind \sys. First, agentic reasoning is often unnecessarily verbose: many turns only require enough internal computation to select the next action, not a full explicit CoT trace. Second, reasoning demand is heterogeneous across turns: while routine turns can be handled with compact latent reasoning, harder turns still benefit from explicit CoT. Rather than uniformly compressing every textual reasoning trace, \sys learns when explicit reasoning is necessary and uses latent reasoning otherwise. This leads to comparable or better EM, much lower token usage, and the best accuracy--efficiency trade-off across both search and tool-use domains.

\subsection{Analysis of Adaptive Mode Selection}

We examine the adaptive mode selection behavior of \sys in the search domain by analyzing per-turn latent and explicit reasoning choices in the 7B evaluation trajectories (\Cref{tab:perturn}).

\stitle{Harder benchmarks retain more explicit reasoning.}
Although latent reasoning dominates overall, \sys uses explicit reasoning more often on harder benchmarks. After AR-GRPO, Bamboogle and 2Wiki have the lowest latent fractions ($59\%$ and $75\%$), while easier single-hop datasets such as NQ and TriviaQA rely on latent reasoning much more frequently ($89\%$ and $91\%$). Compared with the warmed-up policy, AR-GRPO increases latent usage when it is sufficient, as in TriviaQA ($69\%\!\to\!91\%$) and NQ ($84\%\!\to\!89\%$), but keeps it similar or lower on harder datasets such as MuSiQue, 2Wiki, and Bamboogle. This suggests that AR-GRPO learns a task-conditioned mode-selection policy rather than uniformly increasing latent reasoning.

\begin{table}[t]
\centering
\footnotesize
\setlength{\tabcolsep}{4pt}
\begin{tabular}{l ccccc cc}
\toprule
& \multicolumn{5}{c}{Per-turn Latent Fraction} 
& \multicolumn{2}{c}{Total} \\
\cmidrule(lr){2-6} \cmidrule(lr){7-8}
Dataset & T1 & T2 & T3 & T4 & $\geq$5 & Warmup & GRPO \\
\midrule
MuSiQue               & 73 & 85 & 87 & 88 & 78 & 88 & 83 \\
2Wiki                 & 44 & 75 & 84 & 87 & 93 & 75 & 75 \\
HotpotQA              & 56 & 87 & 92 & 94 & 94 & 85 & 84 \\
NQ                    & 72 & 89 & 95 & 95 & 94 & 84 & 89 \\
TriviaQA              & 73 & 95 & 97 & 98 & 95 & 69 & 91 \\
Bamboogle             & 75 & 39 & 49 & 64 & 74 & 68 & 59 \\
\bottomrule
\end{tabular}%
\caption{Per-turn latent fraction on the search domain over all 7B trajectories. Turn-level columns report the latent fraction after AR-GRPO; \emph{Total} compares the overall latent fraction before and after AR-GRPO.}
\label{tab:perturn}
\end{table}

\stitle{Explicit reasoning is concentrated in early planning turns.}
Turn~1 consistently uses the most explicit reasoning, while later turns are mostly latent. This pattern suggests that \sys uses explicit CoT primarily for initial planning, such as decomposing a comparison or compositional question into sub-goals before issuing the first search action. After the initial plan is formed, subsequent retrieve-and-gather turns require less textual deliberation and can usually be handled through latent reasoning. This turn-level behavior supports the design motivation of \sys: explicit reasoning is most useful when the agent must plan or decompose the task, whereas latent reasoning is sufficient for many routine environment-interaction steps.

\begin{table}[t]
\centering
\footnotesize
\setlength{\tabcolsep}{4pt}
\begin{tabular}{c ccc ccc}
\toprule
& \multicolumn{3}{c}{\emph{Search}} 
& \multicolumn{3}{c}{\emph{Tool-use}} \\
\cmidrule(lr){2-4} \cmidrule(lr){5-7}
Latent Steps $K$ 
& EM & Tok & AE 
& EM & Tok & AE \\
\midrule
$1$ & 40.6 & 128 & +0.21 & 87.4 & 109 & +0.90 \\
$2$ & 41.2 & 131 & +0.27 & 88.6 & 114 & +0.94 \\
$4$ & 41.7 & 133 & +0.32 & 89.2 & 117 & +0.95 \\
$8$ & 41.8 & 138 & +0.31 & 89.1 & 124 & +0.94 \\
\bottomrule
\end{tabular}
\caption{Effect of the number of latent steps $K$. Search results are averaged over the search benchmarks using the 7B model, and tool-use results are averaged over the BFCL categories.}
\label{tab:k_ablation}
\vspace{-1em}
\end{table}
\subsection{Effect of Latent Step Length}

\stitle{A few latent steps are sufficient for agentic reasoning.}
We ablate the number of continuous latent steps $K$ used in each latent block in \Cref{tab:k_ablation}. Even with a small number of latent steps, \sys achieves strong accuracy-efficiency trade-offs in both domains. Increasing $K$ from $1$ to $4$ improves EM and AE, showing that the latent mode benefits from a modest amount of internal computation. However, performance changes little beyond $K=4$: using $K=8$ yields negligible accuracy gains while slightly increasing token usage. This suggests that most agentic turns require only compact latent computation to select the next action or incorporate observations, while turns requiring deeper reasoning can still be handled by the explicit mode. We therefore use $K=4$ as the default because it is sufficient to capture most of the benefit of latent reasoning.
\section{Conclusion}

We introduced \sys, an adaptive latent reasoning framework for efficient LLM agents. Instead of merely shortening explicit CoT, \sys changes the reasoning interface: agents use latent reasoning for routine turns and reserve explicit CoT for harder ones. It is trained with Action-Anchored Self-Distillation, which teaches latent reasoning from successful agent actions, and AR-GRPO, which learns when latent reasoning is sufficient for task success. Across search and tool-use domains, \sys achieves comparable or better accuracy with substantially fewer generated tokens than base models and reasoning compression baselines. These findings highlight adaptive use of latent and explicit reasoning as a practical path toward more efficient LLM agents.

\section*{Limitations}

This work has several limitations. First, we focus on agentic tasks that require interaction with an external environment, such as search and tool use. We do not evaluate on domains such as math and coding, which have been heavily used in reasoning post-training and often require less environment interaction. In this paper, we view such settings as closer to single-pass LLM reasoning, where additional reasoning tokens may directly improve final-answer accuracy. By contrast, in agentic settings, more generated reasoning does not necessarily lead to better performance, since many turns mainly require selecting the next environment-coupled action.

Second, \sys relies on successful teacher trajectories for Action-Anchored Self-Distillation, so the latent policy may inherit the coverage and biases of the teacher. Third, we use a fixed latent block length $K$ and a discrete latent/explicit mode choice, leaving more fine-grained control of latent computation to future work. Finally, latent reasoning reduces generated tokens but makes part of the agent's reasoning less interpretable, which may be undesirable when action transparency is important.

\section*{Ethics Statement}

This work follows the ACL Code of Ethics. We use existing public benchmarks and do not collect new human-subject data. The main ethical consideration is that latent reasoning may reduce the transparency of intermediate reasoning compared with explicit CoT. We therefore recommend monitoring agent actions and outputs, and retaining explicit reasoning or additional logging in high-stakes settings.


\bibliography{custom}

\newpage
\appendix

\section{Implementation Details}
\label{app:implementation}
\subsection{Models}
\label{app:models-datasets}

\paragraph{Search domain.}
For the search experiments, we use the released Search-R1 checkpoints based on Qwen2.5:
Search-R1-Qwen2.5-3B\footnote{\url{https://huggingface.co/PeterJinGo/SearchR1-nq_hotpotqa_train-qwen2.5-3b-em-ppo-v0.3}}
and Search-R1-Qwen2.5-7B.\footnote{\url{https://huggingface.co/PeterJinGo/SearchR1-nq_hotpotqa_train-qwen2.5-7b-em-ppo-v0.3}}

\paragraph{Tool-use domain.}
For the tool-use experiments, we use Qwen3-4B-Thinking-2507,\footnote{\url{https://huggingface.co/Qwen/Qwen3-4B-Thinking-2507}}
a reasoning model with native tool-calling capability.

\subsection{Search Environment}
\label{app:search-env}

Following Search-R1, the search agent interleaves reasoning with retrieval over the Wikipedia-18 corpus.
Retrieval uses a FAISS \cite{douze2024faiss} index built on E5-large-v2\footnote{\url{https://huggingface.co/intfloat/e5-large-v2}} embeddings and returns the top-3 documents per query.
Each search trajectory is capped at six turns.
During supervision trajectory construction, we cap each retrieved document at 500 characters and the total context length at 4096 tokens.

\subsection{Latent Block Implementation}
\label{app:latent-block}

Each latent block contains $K=4$ continuous thoughts framed by surface tags \texttt{<LAT>} and \texttt{</LAT>}.
The latent positions are implemented as content-free sentinel placeholders: when the decoder reaches \texttt{<LAT>}, the projector $f_\phi$ writes $K$ continuous embeddings into the subsequent latent positions, and \texttt{</LAT>} is prefilled after the $K$ projections.
The tags are standard vocabulary tokens, so no new special tokens are added.
The projector $f_\phi$ is a two-layer MLP with GELU activation and a final LayerNorm, with hidden width matching the base model.

\subsection{Training Details}
\label{app:training-details}

For the supervised stages, we train for one epoch with LoRA at rank 16 and $\alpha=32$ on the attention q/k/v/o projections, while the projector $f_\phi$ is fully fine-tuned.
We use AdamW with learning rate $1\times10^{-4}$, 3\% linear warmup followed by a cosine schedule, global batch size 32, bf16 precision, and gradient checkpointing.
For AR-GRPO, we continue to use LoRA with the same rank and target modules, while optimizing the projector $f_\phi$ together with the policy.
We use latent bonus coefficient $\alpha=0.3$ in both domains, a KL coefficient of $10^{-3}$ against the SFT reference policy, and length tolerances of $L=400$ for search and $L=1600$ for tool use.

\subsection{Infrastructure}
\label{app:infrastructure}

We use the \texttt{verl} framework with FSDP for the actor and vLLM for rollout generation.
The rollout engine is augmented with a state machine that detects \texttt{<LAT>}, splices projector outputs into the next $K$ sentinel positions, and hot-reloads the projector between rollout steps.

\subsection{Prompts}
\label{app:prompts}

\Cref{tab:system-prompts} specifies the domain-specific system prompts that are used for training and inference. 

\section{Baseline Implementation Details}
\label{app:baselines}

\paragraph{Shared setting.}
All explicit-CoT compression baselines use the same base model, retrieval environment, and evaluation setup as \sys.
Training uses \texttt{verl} with FSDP for the actor and a co-located vLLM rollout engine.
We use LoRA with rank 16 and $\alpha=32$ on the attention q/k/v/o projections, AdamW with learning rate $1\times10^{-6}$, KL coefficient $10^{-3}$ against the Search-R1 actor as the reference policy, a batch of 12 prompts with $G=8$ rollouts per prompt, and 200 update steps in bf16 with gradient checkpointing.

\paragraph{ShorterBetter.}
ShorterBetter uses a group-relative length reward that encourages each rollout to approach the shortest correct reasoning length within its group.
We use EM weight $\alpha=2.0$ and length-penalty weight $\beta=5\times10^{-3}$, computed against the shortest correct rollout among $G=8$ rollouts.

\paragraph{ThinkPrune.}
ThinkPrune applies a staged annealing schedule over the thinking-token budget.
We use three stages with budgets $T\in\{200,120,70\}$, corresponding to the p75, p50, and p25 thinking-length percentiles of the Search-R1 actor's training pool.

\paragraph{O1-Pruner.}
O1-Pruner uses per-prompt reference statistics for length-harmonizing reward computation.
We precompute $(L_{\mathrm{ref}}(x), A_{\mathrm{ref}}(x))$ from $K=8$ rollouts at temperature 0.7 over a 3,200-prompt training sub-pool using the frozen Search-R1 actor.
The online reward uses EM weight $\alpha=2.0$ and clip bounds $(c_{\mathrm{lo}}, c_{\mathrm{hi}})=(-4.0, +2.0)$.
Out-of-pool prompts fall back to the EM-only reward during training.

\section{Use of AI Assistants}
We used Claude Code to assist with implementation and experimentation. We also used ChatGPT to help revise sentences for grammar, clarity, and fluency. 

\section{Licenses}
\begin{itemize}
    \item Natural Questions (NQ): CC BY-SA 3.0
    \item TriviaQA: Apache-2.0
    \item HotpotQA: CC BY-SA 4.0
    \item Wikipedia-18 Corpus: Apache-2.0
    \item MuSiQue: CC BY 4.0
    \item 2WikiMultiHopQA (2Wiki): CC BY-SA 4.0
    \item Bamboogle: MIT
    \item BFCL: CC BY-NC 4.0
    \item ToolMind: Apache-2.0
    \item Search-R1-Qwen2.5-3B: Apache-2.0
    \item Search-R1-Qwen2.5-7B: Apache-2.0
    \item Qwen3-4B-Thinking-2507: Apache-2.0
    \item E5-large-v2: MIT
\end{itemize}

\begin{table*}[t]
\centering
\small
\setlength{\tabcolsep}{5pt}
\renewcommand{\arraystretch}{1.15}
\begin{tabular}{p{0.14\textwidth} p{0.80\textwidth}}
\toprule
\textbf{Domain} & \textbf{System prompt} \\
\midrule
Search &
Answer the given question. You must conduct reasoning first every time you get new information. You may choose either mode per turn: 
\texttt{<latent>••••</latent>} --- compact internal reasoning. Emit exactly four bullet placeholder tokens between the tags; each carries one step of internal latent state. Use by default for routine steps.
\texttt{<think> ... </think>} --- explicit textual reasoning. Use when you need to fuse information from multiple searches, when previous searches were insufficient, or when you need to reflect on your previous reasoning.
After reasoning, if you find you lack some knowledge, you can call a search engine by \texttt{<search> query </search>} and it will return the top searched results between \texttt{<information>} and \texttt{</information>}. You can search as many times as you want. If you find no further external knowledge needed, you can directly provide the answer inside \texttt{<answer>} and \texttt{</answer>}, without detailed illustrations. For example, \texttt{<answer> Beijing </answer>}. \\
\midrule
Tool-use &
You are a careful tool-using assistant. Before each action you must reason. You may choose either mode per turn:
\texttt{<latent>••••</latent>} --- compact internal reasoning. Emit exactly four bullet placeholder tokens between the tags; each carries one step of internal latent state. Use by default for routine steps.
\texttt{<think> ... </think>} --- explicit textual reasoning. Use when you need to chain information from previous tool calls, recover from an unexpected response, or plan a multi-step sequence.
After reasoning, either issue one or more \texttt{<tool\_call>} calls, or produce the final natural-language answer. \\
\bottomrule
\end{tabular}
\caption{System prompts used for the search and tool-use domains.}
\label{tab:system-prompts}
\end{table*}

\end{document}